\documentclass[letterpaper, 10 pt, conference]{ieeeconf}
\usepackage{cite}
\usepackage{amsmath,amssymb,amsfonts} 
\usepackage{graphicx}
\usepackage{textcomp} 
\usepackage{multirow}
\usepackage{gensymb}
\usepackage{xcolor} 
\usepackage[ruled,vlined]{algorithm2e}
\usepackage{gensymb}
\usepackage{textcomp} 
\usepackage{xcolor} 
\usepackage{color}
\usepackage{multirow}
\usepackage{graphicx}
\usepackage{subfigure}
\usepackage[mathscr]{eucal}
\usepackage{array} 

\IEEEoverridecommandlockouts                              

\makeatletter

\makeatletter
\renewcommand{\maketag@@@}[1]{\hbox{\m@th\normalsize\normalfont#1}}%

\setbox0\hbox{$\xdef\scriptratio{\strip@pt\dimexpr
    \numexpr(\sf@size*65536)/\f@size sp}$}

\newcommand{\scriptveryshortarrow}[1][3pt]{{%
    \vcenter{\hbox{\rule[\scriptratio\dimexpr-.2pt\relax]
               {\scriptratio\dimexpr#1\relax}{\scriptratio\dimexpr.4pt\relax}}}%
   \mkern-4mu\hbox{\let\f@size\sf@size\usefont{U}{lasy}{m}{n}\symbol{41}}}}

\makeatother

\makeatletter
\let\NAT@parse\undefined
\makeatother
\usepackage[colorlinks, citecolor=green]{hyperref}

\title{\LARGE \bf
Calib-Anything: Zero-training LiDAR-Camera Extrinsic Calibration Method Using Segment Anything
}

\author{Zhaotong Luo, Guohang Yan$^{\dagger}$ and Yikang Li 
\thanks{$^{\dagger}$ Corresponding author.}
\thanks{Zhaotong Luo , Guohang Yan and Yikang Li are with Autonomous Driving Group, Shanghai Artificial Intelligence Laboratory, China. {\tt\small \{luozhaotong, yanguohang, liyikang\}@pjlab.org.cn}}
}
    
\begin{document}
 
\maketitle

\begin{abstract}
The research on extrinsic calibration between Light Detection and Ranging(LiDAR) and camera are being promoted to a more accurate, automatic and generic manner. Since deep learning has been employed in calibration, the restrictions on the scene are greatly reduced. However, data driven method has the drawback of low transfer-ability. It cannot adapt to dataset variations unless additional training is taken. With the advent of foundation model, this problem can be significantly mitigated. By using the Segment Anything Model(SAM), we propose a novel LiDAR-camera calibration method, which requires zero extra training and adapts to common scenes. With an initial guess, we opimize the extrinsic parameter by maximizing the consistency of points that are projected inside each image mask. The consistency includes three properties of the point cloud: the intensity, normal vector and categories derived from some segmentation methods. The experiments on different dataset have demonstrated the generality and comparable accuracy of our method. The code is available at https://github.com/OpenCalib/CalibAnything.
\end{abstract}

\section{INTRODUCTION}

Camera and LiDAR are the two main types of sensors used in self-driving vehicles. The complementary nature of the two sensors makes them a favored combination in many perception tasks such as depth completion \cite{ma2019self}, object detection \cite{li2022deepfusion} and object tracking \cite{asvadi20163d}. In order to fuse data of these two sensors, calibration is indispensable for time synchronization and spatial alignment. Here we focus on the extrinsic calibration, which is to obtain the transformation matrix between the camera coordinate system and LiDAR coordinate system, including rotation and translation. The accuracy of extrinsic parameters fundamentally limits the result of data fusion. Thus, much effort has been made to handle this problem from different perspectives.




Early methods used artificial targets with special patterns that are easily detected \cite{zhang2004extrinsic, verma2019automatic, xie2018infrastructure, grammatikopoulos2022effective, beltran2022automatic}. It can achieve high precision at the cost of reduced flexibility. Due to the slight drift of the extrinsic parameter in daily use, a more unconditional and automatic method is needed for re-calibration. For this purpose, some methods exploit the geometric features in natural scenes such as lines \cite{kang2020automatic, moghadam2013line, chai2018novel} and vanishing points(VPs) \cite{stamos2008integrating, bai2020lidar}, which often exist in structured scenarios. In order to further remove the constraints of the scene, learning-based approaches take the stage with the assistance of large-scale datasets. It adapts to general scenes and achieves high accuracy. However, simple supervision network\cite{schneider2017regnet} has weak generalization ability and poor interpretability. Although geometric constraints are added \cite{iyer2018calibnet, yuan2020rggnet, lv2021lccnet}, it still needs a large well-labeled dataset and faces accuracy drop under dataset variations.

These problems can be significantly alleviated with the advent of foundation model. The recently released Segment Anything Model(SAM)\cite{kirillov2023segment} is a foundation model for image segmentation, demonstrating an impressive zero-shot ability benefited from the huge amount of training data. Considering the usage of segmentation in calibration \cite{tsaregorodtsev2022extrinsic, rotter2022automatic}, we propose a novel method for LiDAR-camera calibration, without requiring additional training or domain adaptation. 

\begin{figure}[htp]
    \centering
    \includegraphics[width=\linewidth]{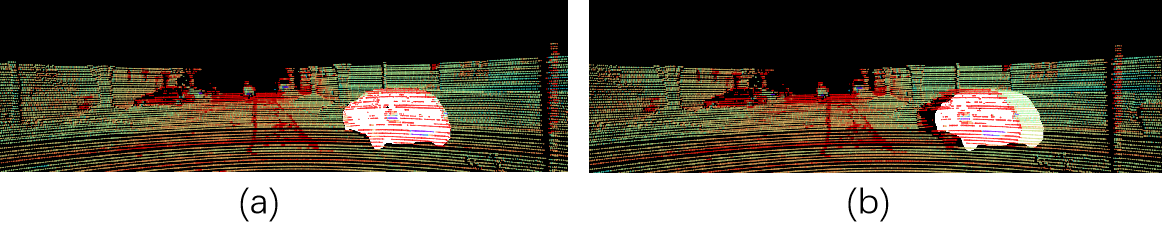}
    \caption{The point cloud projected on a mask of car by the right extrinsic(a) and error extrinsic(b). The color of the points represents intensity value. }
    \label{fig:intensity}
\end{figure}

We first use SAM to perform semantic segmentation on the entire image and get a set of masks. Instead of building definite correspondence between the point cloud and the masks, we calculate the consistency of point cloud attribute
inside the mask, including its intensity, normal vector and segmentation class. As shown in Fig.\ref{fig:intensity}, under the correct extrinsic, the intensity of points inside the car mask have higher consistency. For normal vector, points on the plane masks should have consistent normal direction. The segmentation class of point cloud is obtained simply by plane fitting and euclidean clustering. Objects like vehicles and trunks will be clustered to one class, therefore also has consistency on the masks. We calculate the consistency score of each mask with these three properties. By giving an initial extrinsic, we can optimize it by maximizing the overall score of all masks.

Compared to traditional methods \cite{ma2021crlf}, our method has a higher adaptability to scenarios, as long as there exists orthogonal relationships. Compared to learning-based method \cite{lv2021lccnet}, we don't require extra training on a large well-labeled dataset. Compared to other segmentation-based methods \cite{wang2020soic, tsaregorodtsev2022extrinsic}, we avoid finding determinate correspondences between the image segments and the point cloud, which is always ambiguous or can only be captured in specific types of object. Experiments on different dataset has demonstrated the generality and comparable accuracy of the method.

The contributions of this work is listed as follows: 
\begin{enumerate}
\item We propose a novel automatic LiDAR-camera extrinsic calibration method by using SAM and point cloud consistency that requires zero extra training.
\item The optimizing criteria for extrinsic parameters is the consistency of intensity, normal vector and segmentation class of point cloud on masks, making our method suitable for most scenes. 
\item We valid our method on several dataset to demonstrate its generality and comparable accuracy.
\end{enumerate}

\section{RELATED WORK}
In general, the calibration methods between LiDAR and camera can be divided into target-based and target-free categories. Here we focus more on target-free methods that require little manual work. Traditional ways include using geometric features like lines, maximizing mutual information and ego-motion estimation. By taking advantage of large-scale datasets, learning-based methods are developed to provide a rather accurate calibration result with little scene requirements, roughly divided to regression and segmentation types.

\subsection{Target-Based Methods}
This type of method requires artificial targets, which are always distinctive in color, shape and reflectivity to ease the extraction of features in the image and point cloud. The mainstream of targets are rectangular shaped boards with specific pattern on it, such as checkerboard \cite{zhang2004extrinsic}, circular grid \cite{domhof2019extrinsic} and Apriltag \cite{xie2018infrastructure}. Because the horizontal edge of rectangle may not intersect with LiDAR scans, objects of other shape such as sphere \cite{toth2020automatic}, polygon board\cite{park2014calibration} are also proposed. By building strong correspondences between points-points or points-plane, this type of method generally achieves high precision but needs human intervention at a low or high level.


\subsection{Target-free Methods}
Instead of custom-made target, some methods seek for geometric features in natural scenes. The most used features are lines or edges. There are generally two steps. Firstly the lines in image are detected by an edge detector\cite{moghadam2013line} or segmentation \cite{ma2021crlf}. The lines in the point cloud are mainly obtained by range discontinuity \cite{moghadam2013line,chai2018novel} and intensity difference \cite{ma2021crlf}. Then the many-to-many correspondences between lines are aligned according to its location \cite{moghadam2013line}, intensity and influential range \cite{kang2020automatic}. Besides direct line features, \cite{stamos2008integrating,bai2020lidar} use vanishing points to estimate the rotation matrix. It requires at least two VPs in the scene.

To reduce the dependence on the scene, some methods utilize mutual information to measure the multi-modal registration, including gradient \cite{taylor2015multi}, intensity of the point cloud and the gray value of image \cite{pandey2012automatic, pandey2015automatic}. While above methods requires zones of mutual visibility, motion-based approach estimates the ego-motion of each sensor separately and solve the extrinsic by hand-eye model \cite{ishikawa2018lidar} or minimizing the projection error \cite{park2020spatiotemporal}. The accuracy of calibration is basically limited by the result of visual odometry and LiDAR odometry.

\subsection{Learning-based methods}
The simple paradigm of learning-based methods is using an end-to-end network to estimate the extrinsic parameter with the input of RGB image and depth image. RegNet \cite{schneider2017regnet} first introduces the Convolutional Neural Networks(CNNs) to regress the 6 DoF parameter. In order to improve the generalization ability of the model, geometry constraints are added in the loss function. CalibNet \cite{iyer2018calibnet} trains its network by maximizing the geometric and photometric consistency of the images and point clouds. RGGNet \cite{yuan2020rggnet} considers the Riemannian geometry and utilizes a deep generative model. LCCNet \cite{lv2021lccnet} exploits the cost volume layer for feature matching and predicts the decalibrated deviation from initial calibration to the ground truth. 

Despite of end-to-end networks, learning-based segmentation is used as part of the pipeline. \cite{wang2020soic} performs semantic segmentation respectively on pictures and point clouds, and then matches the centroids of a class of objects in 2D and 3D points. Due to the sparsity of the point cloud, \cite{tsaregorodtsev2022extrinsic} combined multiple frames of LiDAR data together, requiring for a high-precision positioning device. Because of the difficulty of point cloud segmentation, some methods only perform segmentation on the image. \cite{zhu2020online} calibrates the extrinsic by maximizing the number of point cloud fell on the segmented foreground area in image. \cite{rotter2022automatic} uses instance segmentation to obtain object edges and define the loss function by the depth discontinuity. One problem of these methods is that the networks can only predict objects of specific classes, establishing limited correspondences. Besides, a common problem with learning-based methods is the inadaptability to dataset changes.

\section{METHODOLOGY}
\subsection{The Overview of The Method}

\begin{figure*}
    \centering
    \includegraphics[width=\linewidth]{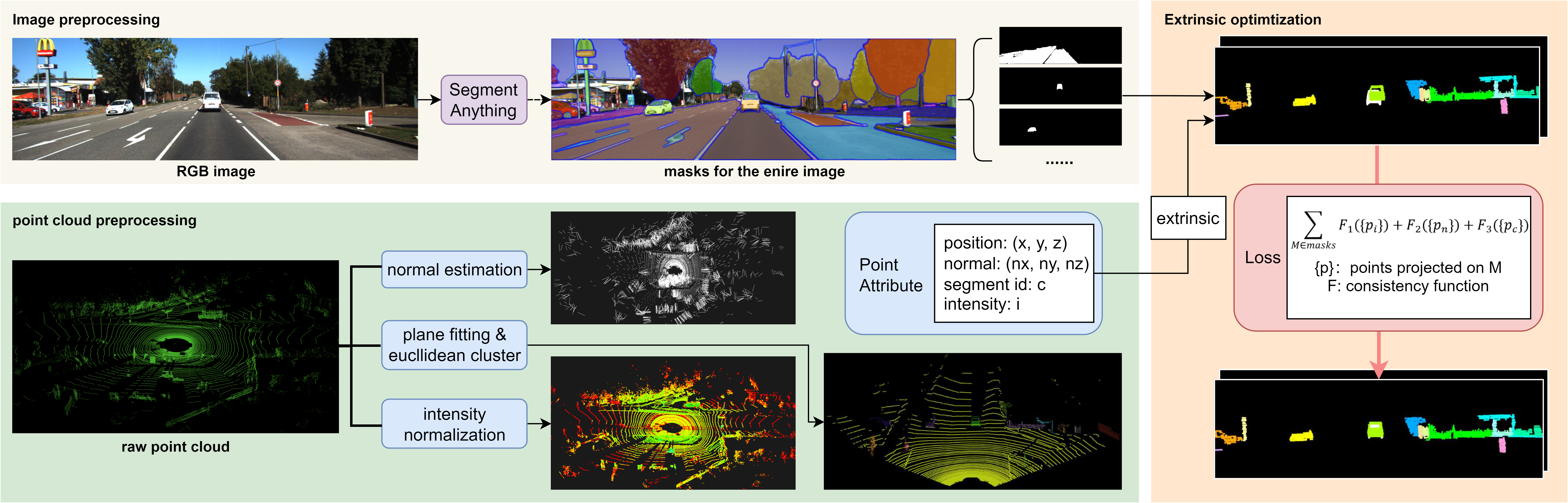}
    \caption{Approach Overview. For image, the Segment Anything Model is used to generate masks of the entire image. For point cloud, we implement normal estimation, simple segmentation methods and intensity normalization to generate corresponding attribute of each point. In the optimization stage, the point cloud is projected to the masks by the extrinsic. We design a loss function that is determined by the attribute consistency of the points inside the mask area.}
    \label{fig:pipeline}
\end{figure*}

The whole process can be divided into three parts. For image segmentation, we use SAM to generate the masks of the entire image. For point cloud, we implement normal estimation, simple segmentation methods and intensity normalization to generate corresponding attribute of each point. Then the optimization target is to make the points that fall on one mask has close attribute value. We have designed a score function to evaluate the consistency. Several rounds of search are performed to obtain the final result. Fig.\ref{fig:pipeline} shows the pipeline of our proposed method.

\subsection{Data Pre-processing}
\subsubsection{Image Segmentation}
\

SAM is first applied on the entire image to get a number of masks of differentiated objects. Since we use the consistency of the point cloud, we want the segmentation to be more fine-grained and detailed. Thus we adjust the hyper parameter of SAM to obtain more masks with less overlapping areas. The masks are annotated as $\mathbb{M} = \{M_i\ |\ i=0,1,..., N\}$. Each mask is a binary matrix of the same size as the image. The value $M_i(u,v)= \{0 ,1\}$ denotes whether the pixel $(u, v)$ belongs to segment i.

\subsubsection{Point cloud Preprocessing}
\ 

There are three parts of preprocessing: normal estimation, intensity normalization and segmentation.

For normal estimation, there are a number methods \cite{rusu2010semantic, boulch2012fast} that can be directly used. Here we choose a simple one which is enough for our application. The normal direction of a point on the surface is approximated as the normal of a plane tangent to the surface. The plane normal can be estimated by the analysis of the eigenvectors and eigenvalues or Principal Component Analysis(PCA) of a covariance matrix created from a number of nearest neighbors of the query point. The K-d tree data structure is used for efficient k-nearest-neighbors(KNN) search. 

The intensity of the point cloud is normalized by a scale factor for subsequent consistency calculation in the case that point cloud intensity ranges different according to the LiDAR type. 

Besides these two attributes, we conduct simple segmentation methods to the point cloud. We first apply plane fitting by RANSAC algorithm to extract large planes in the scene, such as the ground and walls. Then we apply some euclidean clustering \cite{rusu2010semantic} to the remained point cloud and get clusters of individual objects like vehicles and trees. We assign a number $c$ to the point, indicating which class it belongs to.

The final attribute of a point in the point cloud can be represented as:
\begin{equation}
    p = \{x, y, z, n_x, n_y, n_z, r, c \}
\end{equation}
which is the position, normal vector, reflectivity and segmentation class of the point P.

\subsection{Extrinsic Optimization}
\subsubsection{Consistency function}
\

The point p can be projected to the image frame by an initial extrinsic $T$:

\begin{equation}
    \lambda \left ( \begin{matrix}p_u \\ p_v \\ 1\end{matrix} \right ) = K T \left ( \begin{matrix} p_x \\ p_y \\ p_z \\ 1 \end{matrix} \right )
\end{equation}
here we assume the intrinsic $K$ is already known. 

Then for each mask $M_i$, we can get a set of points falls on it:

\begin{equation}
    P_i = \{p \ |\ M_i(p_u,p_v)=1\}
\end{equation}

The score measuring the consistency of the points set $P_i$ can be calculated as:

\begin{equation}
    s_i = (w_R F_R(P_i) + w_N F_N(P_i) + w_S  F_S(P_i))f(||P_i||)
\end{equation}
where $F_R(\cdot ),\ F_N(\cdot ),\ F_S(\cdot )$ is the corresponding function of reflectivity, normal vector and segmentation class. $w_R,\ w_N,\ w_S$ is the weight for them. In practice we use $w_R = w_N = w_S$. And $f(\cdot )$ is the adjusting function according to the number of points in $P_i$.

The reflectivity consistency is simply calculated by the Standard Deviation(std) of all values:
\begin{equation}
    F_I(P_i) = 1 - \frac{1}{n}\sum\limits_{p\in P_i}{(p_r - \overline{p_r})^2}
\end{equation}

Suppose matrix A of size $(3\times n)$ is composed by normal vectors in $P_i$. The consistency function $F_N$ is represented as:
\begin{equation}
    F_N(P_i) = \frac{1}{n^2} \sum_{v\in A^TA} |v|
\end{equation}
It is the average of the pairwise dot product of all vectors.

For segmentation class, the points of each category are first counted and sorted from largest to smallest. This is denoted as $(c_0, c_1, ...)$, where $c_i$ is the number of points in the i-th largest class. The consistency is the weighted sum of all classes:

\begin{equation}
    F_S(P_i) = \frac{1}{C}\sum_i{ k^ic_i},(i=0,1,...) 
\end{equation}
\begin{equation}
    C = \sum_i{c_i}, (i=0,1,...)
\end{equation}
where k is a scale factor, using $k=0.4$ in practice.

\begin{figure}[htp]
    \centering
    \includegraphics[width=0.7\linewidth]{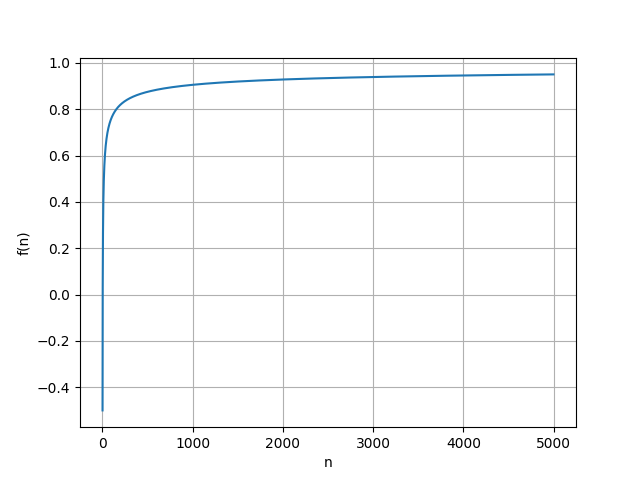}
    \caption{The adjust function is a monotonically increasing function. When $n \in [1,+\infty), f(n) \in (0, 1) $ }
    \label{fig:funciton}
\end{figure}

The adjusting function is used to compensate for the loss of consistency caused by a larger point number. It is represented as:
\begin{equation}
    f(n) = 1 - k_1 n^{k_2}, (k_1 > 0, k_2 < 0)
\end{equation}
where n is the number of points in $P_i$. $k_1$ and $k_2$ are empirically set to 1.5 and -0.4. The function curve is shown in Fig.\ref{fig:funciton}. The set with a small number of points will be punished by a lower coefficient.

The final consistency score is the weighted mean of scores of all masks:
\begin{equation}
    s = \sum_{i} {w_i s_i},\ w_i = \frac{||P_i||}{\sum_i{||P_i||}}
\end{equation}
the weight of each mask is set according to the number of points that projected on the mask.

\subsubsection{Extrinsic Search}
\

For each extrinsic, we can calculate the consistency score to evaluate the alignment between the image and the point cloud. Then we can search for the best extrinsic with an initial guess of it. There are two steps: first we use Brute force search with large strides, only calibrating its rotation. Then we apply random search in a smaller range to refine both rotation and translation. 

Because the translation error is often small and has little effect on the projection, we only change rotation in the brute-force search phase. For an initial guess $T_{init}$, we first calculate its score $s_{init}$. Then we uniformly sample the increment of 3 DoF rotation parameters in the range $[-A, A]$ degree near the initial value, with stride s. If a higher score is achieved, the extrinsic will be updated. In the refined phase, the increment of the 6 DoF extrinsic parameter is randomly sampled in a smaller range with fixed number of times.

\section{EXPERIMENTS}
\vspace{-1mm}
\subsection{Experiment Settings}
We conduct experiment on two datasets. The first one is based on KITTI odometry benchmark \cite{geiger2013vision}. The second one is our own dataset collected by a HESAI Pandar64 LiDAR and a color camera(FOV=60$\degree$). 

\subsection{Qualitative Results}
In this part, we visualize the projection before and after the calibration to qualitatively show the effect of our method.

\begin{figure}[htp]
    \centering
    \includegraphics[width=\linewidth]{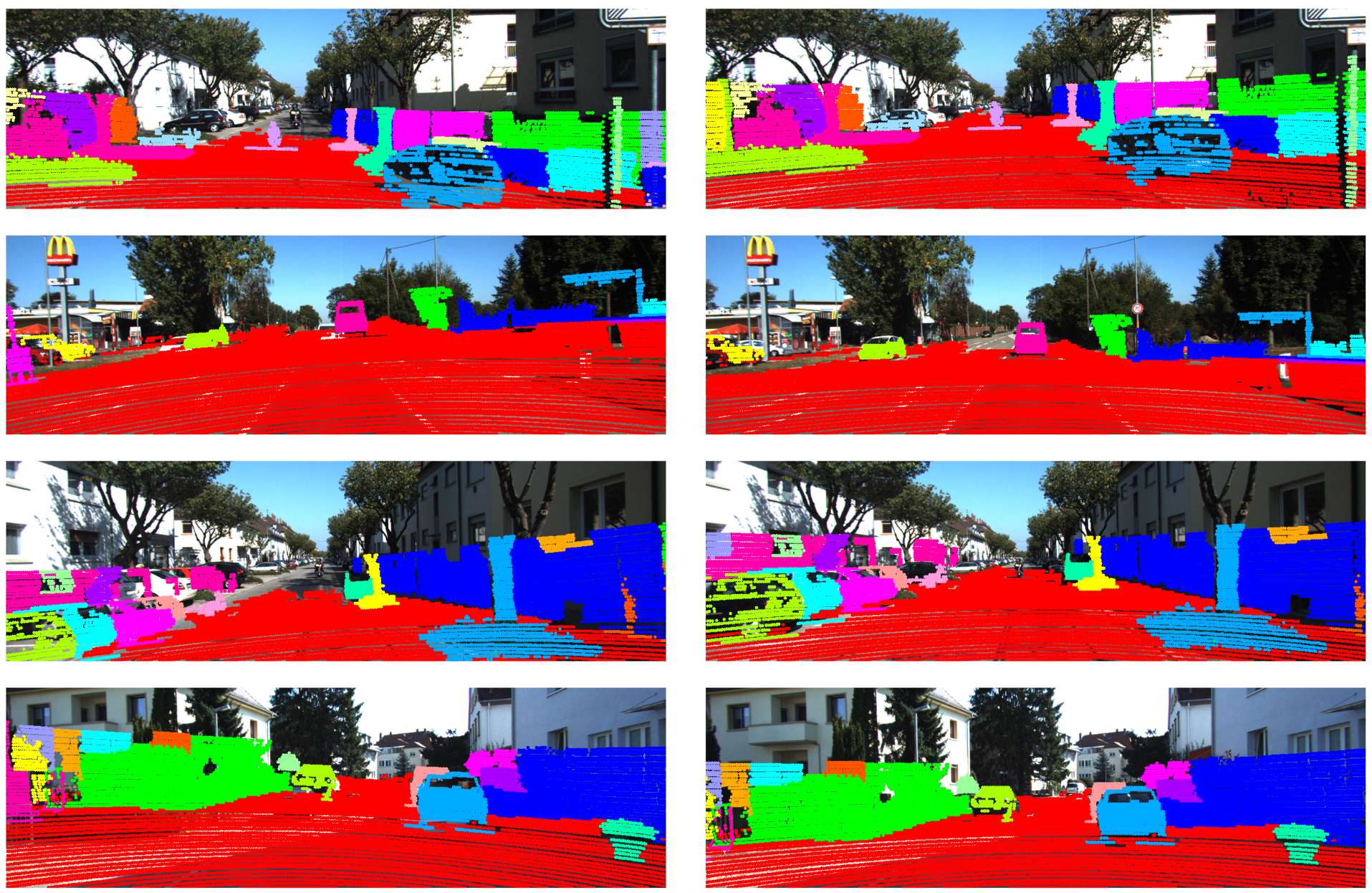}
    \caption{The projection before and after calibration. The initial extrinsic has an error around 3 degree.}
    \label{fig:calibration}
\end{figure}

Given an initial extrinsic parameter with an error of about 5 degrees, the algorithm can correct it back to the right projection, as shown in Fig.\ref{fig:calibration}.

\subsection{Quantitative Results}
We compare the accuracy of our method with other segmentation-based methods in the metric of L2 loss and Huber loss. L2 loss is the mean vector norm of the displacement error vector of translation $\overline{\|  \triangle \textbf{t}\|}$ and rotation $\overline{\|  \triangle \alpha \| }$. Huber loss \cite{huber1992robust} is less sensitive to outliers.

\renewcommand\arraystretch{1.5}

\begin{table}[htp]
    \centering
    \caption{Quantitative evalution results}
    \begin{tabular}{c|c|c|c|c|c}
         \hline
         \multirow{2}*{Method} & \multirow{2}*{Dataset} & \multicolumn{2}{c}{L2 Loss} & \multicolumn{2}{|c}{Huber Loss} \\
         \cline{3-6}
         ~ & ~ & $\overline{\|  \triangle \textbf{t}\|}$ & $\overline{\|  \triangle \alpha \| }$ & $\overline{\|  \triangle \textbf{t}\|}$ & $\overline{\|  \triangle \alpha \| }$ \\
         \hline
         \cite{tsaregorodtsev2022extrinsic} & KITTI & 20.2cm & 0.34$\degree$ & 20.3cm & 0.33$\degree$ \\
         \hline
         ours & KITTI & 10.7cm & 0.174$\degree$ & 10.4cm & 0.168$\degree$ \\
         \hline
         ours & ours & 12.8cm & 0.203$\degree$ & 12.6cm & 0.200$\degree$ \\
         \hline
    \end{tabular}
    \label{accuracy}
\end{table}


\section{CONCLUSIONS}
In conclusion, we propose a novel LiDAR-camera calibration method using Segment Anything and point cloud consistency. Our approach can adapt to infrastructure scenarios without requiring additional training on a well-labeled dataset. In the future, we will conduct experiment on more datasets to prove the generality of our methods. Quantitative experiments on other methods will also be implemented to prove our comparable accuracy. Since the ground truth of the extrinsic parameters may also have errors, we will further use the stability of the estimate as another evaluation metric.

\bibliographystyle{IEEEtran}
\bibliography{egbib}

\end{document}